%% file: cnnlevset.tex
\documentclass[runningheads,a4paper]{llncs2e/llncs}

\usepackage{amssymb}
\usepackage{graphicx}
\graphicspath{{files/fig/}}

\usepackage{algorithm}
\usepackage[noend]{algpseudocode}

\usepackage{url}
\urldef{\mail}\path|min.tang@ualberta.ca|
\newcommand{\keywords}[1]{\par\addvspace\baselineskip
\noindent\keywordname\enspace\ignorespaces#1}

\usepackage{enumitem}

\makeatletter
\def\BState{\State\hskip-\ALG@thistlm}
\makeatother
\begin{document}

\mainmatter  % start of an individual contribution

\title{A Deep Level Set Method for Image Segmentation}

\titlerunning{A Deep Level Set Method for Image Segmentation}
% A discriminative deep levelset method for image segmentation 


% figure out the order ; we might have to add Alan Wilman
\author{Min Tang \and Sepehr Valipour \and Zichen Zhang \and \\
Dana Cobzas \and Martin Jagersand}



\institute{Department of Computing Science, University of Alberta, Edmonton, Canada \\
\mail\\}
% \institute{-}
\authorrunning{M. Tang \and S. Valipour \and Z. Zhang \and D. Cobzas \and M. Jagersand}
\toctitle{Lecture Notes in Computer Science}
\tocauthor{Authors' Instructions}
\maketitle

\begin{abstract}
\input{files/abstract}
\end{abstract}

\section{Introduction}
\input{files/introduction}

\section{Methods}
\input{files/method}
\section{Experiments and Results}
\input{files/experiments}

\section{Discussion}
\input{files/discussion}

%--------------------
\bibliographystyle{ieeetr}
\bibliographystyle{llncs2e/splncs03}
\bibliography{segm}

\end{document}

%% file: files/abstract.tex
%Deep architectures, in particular Fully Convolutional Networks (FCN) have recently been shown to be quite succesful at segmentation tasks. Nevertheless, FCNs have two main limitations compared to traditional energy-based approaches: they have no explicit way of incorporating smoothing and prior information (e.g. shape), and the network requires a lot of training data which may be expensive to obtain in medical imaging. To overcome these limitations, we propose a novel image segmentation approach that integrates FCN with a levelset model. The levelset model is not just a post-processing tool but is integrated into the training to refine the FCN. This technique allows the use of unlabeled data during training like in semi-supervised learning. Using two types of medical imaging data (liver CT and heart MRI data), we show that the integrated model achieves good performance even when little training data is available, outperforming the original FCN or levelset alone. In addition, the unified model trains the FCN such that it can perform quite well on its own at the end of training. 

%In addition, the unified iterative model trains the FCN such that it can achieve comparable performance with an FCN trained using twice or three times as much data.

This paper proposes a novel image segmentation approach that integrates fully convolutional networks (FCNs) with a level set model. Compared with a FCN, the integrated method can incorporate smoothing and prior information to achieve an accurate segmentation. Furthermore, different than using the level set model as a post-processing tool, we integrate it into the training phase to fine-tune the FCN. 
This allows the use of unlabeled data during training in a semi-supervised setting.  
Using two types of medical imaging data (liver CT and left ventricle MRI data), we show that the integrated method achieves good performance even when little training data is available, outperforming the FCN or the level set model alone. 

%% file: files/introduction.tex
Image segmentation is one of the central problems in medical imaging. It is often more challenging than natural image segmentation as the results are expected to be highly accurate, but at the same time, little training data is provided.

To address these issues, often strong assumptions and anatomical priors are imposed on the expected segmentation results.
For quite a few years, the field was dominated by energy-based approaches, where the segmentation task is formulated as an energy minimization problem. Different types of regularizers and priors \cite{Cremers-et-al-ijcv07} can be easily incorporated into such formulations.
%Most traditional methods utilize hand-crafted features incorporated into an energy-based segmentation method or a classifier. 
% For quite a few years, the field was dominated by energy-based approaches, where the segmentation task is formulated as an energy minimization problem either in a continuous domain \cite{Caselles97geodesics,ChanVese01regions} or discretized on a graph \cite{Boykov06GC}. Different types of regularizers and priors \cite{Cremers-et-al-ijcv07} can be easily incorporated into such formulations, making the optimization problem easier and the solution closer to the target. Also, principled solutions to solve such problems have been proposed using calculus of variations and PDEs \cite{Caselles97geodesics,ChanVese01regions} or convex optimization \cite{Boykov06GC,chan_convexICIP05}. 
Since the seminal work of Chan and Vese \cite{ChanVese01regions}, the level set has been one of the preferred models due to its ability to handle topological changes of the segmentation function. Nevertheless, there are some limitations of traditional level set approaches: they rely on a good contour initialization and a good guess of parameters involved in the model. Additionally, they often have a relatively simple appearance model despite some progress \cite{Cremers-et-al-ijcv07,ismailTMI10}.

%Machine learning and in particular 
The recently introduced deep neural net architectures address some of these issues by automatically learning appearance models from a large annotated dataset. Moreover, FCNs \cite{longFCN15} have been proved successful in many segmentation tasks including medical imaging \cite{unet2D2015,broschMICCAI15}. 
% Convolutional neural networks (CNN) pose the segmentation problem as voxel-wise classification and require thousands of inferences for a testing image. To address this problem, fully convolutional networks (FCN) \cite{longFCN15} replace the first several fully-connected layers of a standard CNN with convolutional layers. 
% FCN are computationally efficient and have proved very successful in many segmentation tasks including medical imaging \cite{unet2D2015,broschMICCAI15}. 
Despite their success, FCNs have a few limitations compared to traditional energy-based approaches: they have no explicit way of incorporating regularization and prior information. The network often requires a lot of training data and tends to produce low-resolution results due to subsampling in the strided convolutional and pooling layers.

We address these limitations by proposing an integrated FCN-levelset model that iteratively refines the FCN using a level set module. We show that (1) the integrated model achieves good performance even when little training data is available, outperforming the FCN or level set alone and 
(2) the unified iterative model trains the FCN in a semi-supervised way, which allows an efficient use of the unlabeled data. In particular, we show that using only a subset of the training data with labels%(30$\%$ of training data in the liver segmentation; 50$\%$ in the ventricle segmentation)
, the jointly-trained FCN achieves comparable performance with the FCN trained with the whole training set.  

Few other works address the problem of introducing smoothness into convolutional nets by using explicit regularization terms in the cost function \cite{ghassan_cnntopol_miccai16} or by using a conditional random field (CRF) either as a post-processing step \cite{deepmedMIA17,cai_cnncrf_miccai16} or jointly trained \cite{zheng_crfcnnICCV15} with the FCN or CNN. However, only specific graphical models can be trained with the FCN pipeline, and they cannot easily integrate shape priors. A joint deep learning and level set approach has also been recently proposed \cite{Ngo_cnnlevset_mia17,chen2013deep}, but their work considers a generative model (deep belief network DBM) that is not trained by the joint model. 
%Our proposed joint FCN-level set model not only produces good segmentation results with little training data but is also capable of further improving the network itself through integrated training. 
% comment on the low res - skip architectures (unet) or shallow networks (brosch15) 

%% file: files/method.tex
\begin{figure}[t]
\centering
\includegraphics[width=0.79\linewidth]{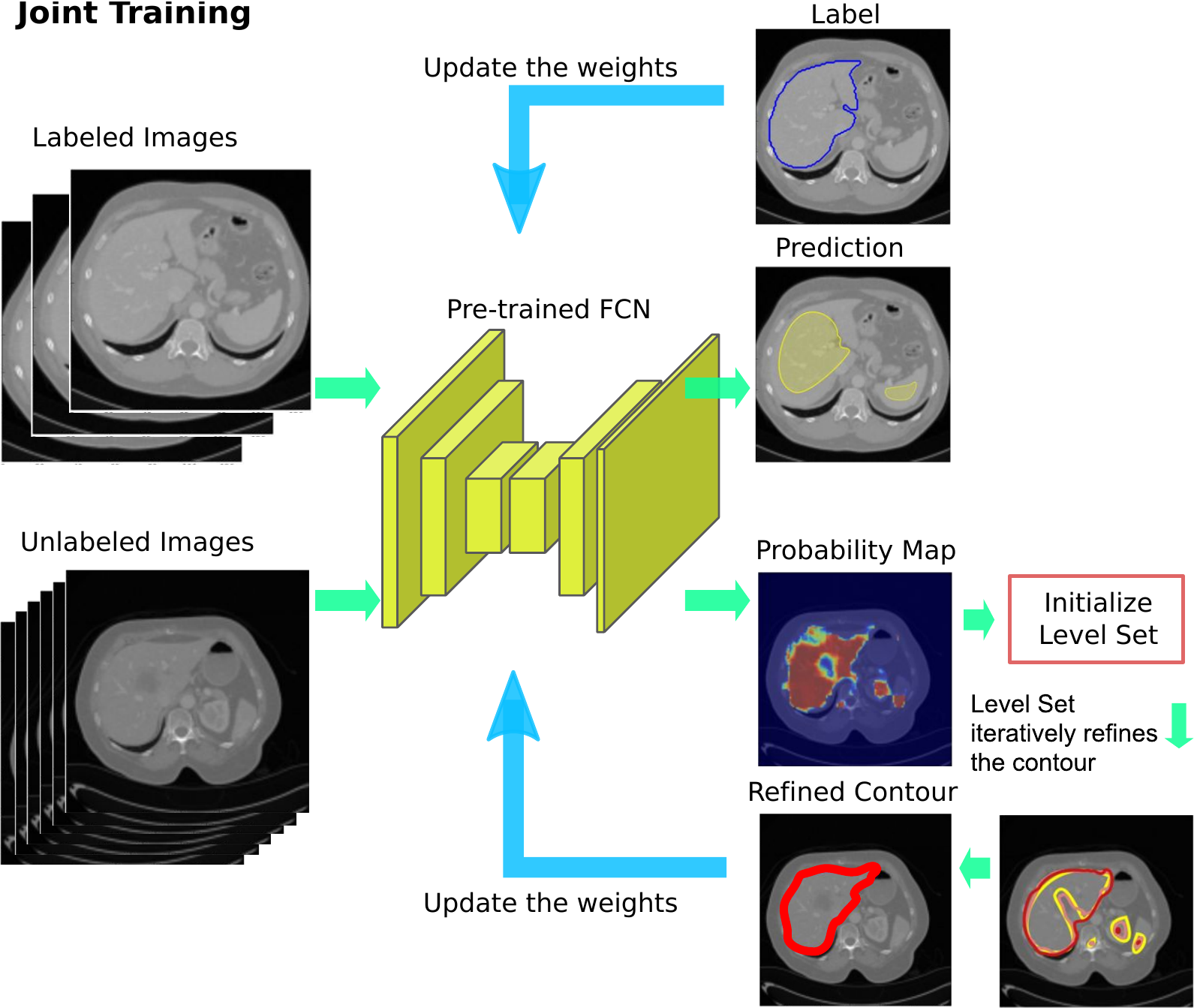}
\caption{Overview of the proposed FCN-levelset model. The pre-trained FCN is refined by further training with both labeled (top) and unlabeled data (bottom). The level set gets initialized with the probability map produced by the pre-trained FCN and provides a refined contour for fine-tuning the FCN. Green and blue arrows denote the forward pass and the back-propagation, respectively.}
%Level set gets initialized with the probability map produced by FCN and provides a refined contour for fine-tuning the FCN. This integrated model allows the use of  unlabeled images for training FCN. The joint training is performed with both labeled and unlabeled images to prevent drastic changes to the network weights.}
\label{fig:sys}
\end{figure}

An overview of the proposed FCN-levelset framework is shown in Fig. \ref{fig:sys}.
The FCN is pre-trained with a small dataset with labels. Next, in the semi-supervised training stage, the integrated FCN and level set model is trained with both labeled (top half of  Fig.\ref{fig:sys}) and unlabeled (bottom half of Fig.\ref{fig:sys}) data. The segmentation is gradually refined by the level set at each iteration. Based on the refined segmentation the loss is computed and back-propagated through the network to improve the FCN. With the FCN trained in this manner, inference is done in the traditional way. A new image is fed to the network and a probability map of the segmentation is obtained.  The output can be further refined by the level set if desired. The following subsections show details on the level set model as well as the integration with FCN.
%Then it is combined with the level set for semi-supervised training using unlabeled data. 
%During this procedure, FCN runs a forward pass to get a coarse prediction and a probability map for the positive class. This probability map is used by level set as an initialization mask. The level set refines this segmentation slightly at each iteration. The loss is computed based on the refined segmentation and back-propagated through the network to improve the FCN. 
%The next three subsections show details on the level set model and the FCN components as well as their integration. 

\subsection{The level set method}
Following traditional level set formulations \cite{ChanVese01regions,Paragios02geodesic}, an optimal segmentation is found by minimizing a functional of the following form:
\begin{equation}\label{eq:1}
E(\Phi, \mathcal{A}) = E_{shape}(\Phi, \mathcal{A}) + \mu E_{smooth}(\Phi) + \lambda E_{data}(\Phi)
\end{equation}
where $\Phi(\cdot)$, defined over the image domain $\mathbf\Omega$, is a signed distance function that encodes the segmentation boundary. 
$\mu$ and $\lambda$ are constants manually tuned and kept fixed during experiments.
The nonuniform smoothness term has the form of a weighted curve length:
\begin{equation}\label{eq:smooth}
   E_{smooth}(\Phi) = \int\int_{\mathbf\Omega}\delta_\alpha(\Phi(x))b(\cdot)|\nabla\Phi(x)|\ d\mathbf\Omega, 
\end{equation}
where $\delta_\alpha(\cdot)$ is the regularized Dirac function. 
The weights $b(\cdot)$ are inversely proportional to the image gradients.  
The data term models the object/background intensity distribution as:
\begin{equation} \label{eq:data}
E_{data}(\Phi) =\int\int_\mathbf\Omega H_\alpha(\Phi(x))\log(P_O(x))d\mathbf\Omega + \int\int_\mathbf\Omega (1 - H_\alpha(\Phi(x)))\log(P_B(x))d\mathbf\Omega
\end{equation}
where $H_\alpha(\cdot)$ is the Heaviside function. $P_O: \mathcal{R^+} \rightarrow [0,1]$ and $P_B: \mathcal{R^+} \rightarrow [0,1]$ are the probabilities belonging to object/background regions. In our model, the probabilities estimated by the FCN are used. 
The shape term is a critical component in knowledge-based segmentation.
Based on the squared difference between the evolving level set and the shape prior level set, we choose \cite{cremers_shapeIJCV06}:
\begin{equation}
E_{shape}(\Phi) = \int\int_\mathbf\Omega \delta_\alpha(\Phi(x))\big(S\Phi(x) - \Phi(x)_\mathcal{M}(\mathcal{A}(x))\big)^2d\mathbf\Omega
\end{equation}
where $\Phi(\cdot)_\mathcal{M}$ denotes the shape model.
${\mathcal A(x)} = S\mathbf{R}x + \mathbf{T}$ is an affine transformation between the shape prior and the current segmentation. 

To optimize the segmentation energy functional Eq. (\ref{eq:1}), calculus of variations is utilized. The position of the curve is updated by a geometric flow.
% \begin{equation}
% \frac{\partial}{\partial\tau}\Phi(x) = - \frac{d}{d\Phi(x)} E_{shape}(\Phi(x), C) - \mu\frac{d}{d\Phi(x)} E_{smooth}(\Phi(x)) - \lambda\frac{d}{d\Phi(x)} E_{data}(\Phi(x))
% \end{equation}
Jointly with the evolution of $\Phi$, the transformation parameters ${\mathcal A}$ are also estimated.

\subsection{The integrated FCN-levelset model}
\label{sec:fcn-levelset}
%The level set method is sensitive to the initialization which sets a limitation for the generalization. Different initialization conditions lead to different and often completely wrong segmentations. To avoid this issue, level set methods are commonly used in conjunction with an initial manual user segmentation. In addition, traditional level set models have simplistic intensity models that do not account for large variability in the data. 
%The way to get a automatic way to get the initialization is important for level set.
%On the other hand, FCN based segmentation can be applied on raw images with no initialization and are able to learn quite complex intensity models. But FCNs can not easily incorporate the prior knowledge (e.g. smoothing, shape priors). 
%They also have trouble getting exact contours as the pooling layers strip them from such precision.

{\bf FCN.}
We create a shallow FCN to make the network less prone to over-fitting on a small training set. 
Fewer pooling layers are added to achieve a finer segmentation. 
% We only have 4 convolutional layers for liver segmentation, while 7 convolutional layers for the left ventricle segmentation.
We only have 4 convolutional layers with 23938 parameters and a total subsampling rate of 8 for liver segmentation, while 7 convolutional layers with 51238 parameters and a total subsampling rate of 6 are used for the left ventricle segmentation.
Each convolution contains a $3\times3\times3$ filter followed by a rectified linear unit (ReLu), and then a max pooling with strides of two. An upconvolution is added right after a $1 \times1\times1$ convolution to achieve a dense prediction.
During training, 
we randomly initialize all new layers by drawing weights from a zero-mean Gaussian distribution with standard deviation 0.01 and ADADELTA\cite{zeiler2012adadelta} was used to optimize the cross entropy loss:
\begin{equation} \label{eq:loss}
L(y,x;\theta) = -\frac{1}{N}\sum_{i=1}^{N} [ y^{(i)}\ln P(x^{(i)};\theta) + (1 - y^{(i)})\ln (1 - P(x^{(i)};\theta)) ]
\end{equation}
where $N$ is the number of pixels in one image, ($x^{(i)}$,$y^{(i)}$) denotes the pixel $i$ and its label, $\theta$ is the network parameter and $P(x^{(i)};\theta)$ denotes the network predicted probability of $x^{(i)}$ belonging to the object.

\noindent
{\bf FCN-levelset.} 
%image size input and output, learning rate
The pre-trained FCN is further refined by the integrated level set module. %as illustrated in Fig. \ref{fig:sys}. 
Each unlabeled image is fed to the FCN and produces a probability map.
It provides the level set module with a reliable initialization and foreground/background distribution for $E_{data}(\Phi)$ in Eq. (\ref{eq:data}). 
The level set further refines the output of the FCN.
We compute the cross entropy loss between the FCN prediction and the level set output as the label. This loss is back propagated through the network to update the weights $\theta$. 
In this manner, the FCN can implicitly learn the prior knowledge which is imposed on the level set especially from the unlabeled portion of dataset.
%The cross entropy loss is computed based on the FCN prediction and the refined output as label and back propagate through the network to update the weights $\theta$.
%the refined output of the level set is back propagated to tune the FCN weights.
%The loss is computed by using the refined output and back propagate through the network and the weights $\theta$ are updated.
Tuning the model weights only with the unlabeled data may cause drastic changes and corrupt the well learned weights $\theta$ from the labeled data. 
This is especially important at the beginning of the joint training when the performance of the system is not yet good.
To make the learning progress smooth, the integrated FCN-levelset model is trained with both labeled and unlabeled data as illustrated in Fig. \ref{fig:sys}. 
%$\{(X_i, Y_i, {X_u}_j); i = 1, 2, \cdots, N, j = 1, 2, \cdots, M\}$. $(X_i, Y_i)$ and ${X_u}_j$ denote the labeled data and unlabeled data, respectively. 
%The probability map obtained from the pre-trained FCN 
%on the unlabeled data that 
%provides the level set module with a reliable initialization and foreground/background distribution for $E_{data}(\Phi(x))$ in Eq. (\ref{eq:data}). 
%The level set further refines the output of the FCN. %Unlike pure post-processing, the refined output of the level set is back propagated to tune the FCN weights. 
% through $({X_u}_j, {Y_u}_j)$. 
%Fig. \ref{fig:sys} illustrates the process.
%Note that, this integration is unlike a simple post processing, as we use the level set-refined segmentation of the unlabeled data as labels for the FCN and train it with them in a semi-supervised fashion.
% Algorithm \ref{alg1} delineates the training procedure. $X,Y$ are images and the corresponding labels, while $Xu$ are unlabeled train images. $F$ is the FCN network and $L$ is the cross-entropy loss function Eq. (\ref{eq:loss}). $MAX\_BUFFER$ is the capacity of the memory buffer and $K$ is the batch size.

During the joint training, to ensure a stable improvement, the memory replay technique is used. It prevents some outliers from disrupting the training. In this technique, a dynamic buffer is used to cache the recently trained samples. Whenever the buffer is capped, a training on the buffer data is triggered which updates the network weights. Then the oldest sample is removed from the buffer and next training iteration is initiated.
% Min: maybe should detail the buffer size???
% First, the FCN network is trained by back propagating the loss of mini-batches taken from labeled portion of the train data (lines 2 to 5). Then, the joint training starts where the goal is to use refined level set output of unlabeled data as weak labels for re-training the network. However, re-training with only these labels may corrupt the learned weights of the pre-training phase. It is especially important at the beginning of the joint training when the network performance is low and therefore the weak labels are not close to the ground truth. During the joint training we use both labeled data and unlabeled data to ensure stable and smooth improvement. In line 8 of the algorithm, mini-batches are chosen from both $X,Y$ and $Xu$. We also use a memory replay technique to prevent some outliers disrupting the training. In this technique, a dynamic buffer collects mini-batches (line 10). When there are enough samples in the buffer ($MAX\_BUFFER$), a training on the buffer data is triggered (lines 12 to 14) which updates the network weights. Finally, $K$ oldest samples are removed from the buffer and we move to the next iteration.

\noindent
{\bf Inference.}
%{\bf Testing} 
To infer the segmentation, a forward pass through the FCN is required to get the probability map. The level set is initialized and the data term is set according to this probability map. Then the final segmentation is obtained by refining the output contour from the FCN. Different from the training where level set is mainly used to improve the performance of the FCN, inference is a post-processing step and the refined output is not back propagated to the FCN.  

%% file: files/experiments.tex
%datasets: liver , tumour , heart (?)  
%
%evalation metrics
%
%comparative results that show how the integrated approach (4) outperforms the other combinations (1-3)
%(1) FCN
%(2) levset
%(3) FCN + levset (postprocessing)
%(4) FCN-levset 
%image + table/bargraphs
%
%show that after training FCN performs quiet well on its own
%
%shape prior - even if not full experiments an evidence of its advantage into such a system would be really good. 
%
%params to discuss - how they influence the system (??)
%- number of training samples

%[dana] 
% put some subsection in experiments
% 1 implementation
% 2 datasets
% 3 results : 

\subsection{Data}

%[dana] mention that our system is 2D and all experients are 2D for now.
%[vincent] added to the top of the experiment section
\noindent
\textbf{Liver Segmentation} contains a total of 20 CT scans \cite{van20073d}. 
%Most of them were pathologic and included tumors, metastasis and cysts in different sizes. 
All segmentations were created manually by radiology experts, working slice-by-slice in transversal view. The scans are randomly shuffled and divided into training (10 scans), validation (5 scans) and testing (5 scans). We select the middle 6 slices from each scan and form our data sets.

\noindent
\textbf{Left Ventricle Segmentation} is a collection of 45 cine-MR sequences (about 20 slices each) taken during one breath hold cycle \cite{lv_challenge}. Manual segmentation is provided by an expert for images at end-diastolic (ED) and end-systolic (ES) phases. The original data division was used in our experiments where the 45 sequences are randomly divided into 3 subsets, each with 15 sequences, for training, validation, and testing. All slices are selected from the scans.   
% [dana] we need test train volumes and how many slices selected for 2D approach
%We cropped edges of input images for 64 pixels from each side as the ROI solely lies in the center of images. Images are also mean normalized before being used for training. 

For both datasets, we divide the training set in two, labeled (30\% of the liver data; 50\% of the left ventricle data) and unlabeled part. The ground truth segmentation is not provided for the unlabeled part to simulate a scenario where only a percentage of data is available for supervised training. 

\subsection{Experiments}
\label{sec:exp}
% Min: input image size: original size??? 
The networks used in the experiments are described in Section \ref{sec:fcn-levelset}.
As the level set method is sensitive to the initialization, the FCN needs to be pre-trained on the labeled part of data. During training, the batch size was 12, and the maximum number of epochs was set to 500. The best model on the validation set was stored and used for evaluations.

The level set models used for liver and left ventricle data are slightly different, depending on their properties. For liver data, we use a weighted curve length as the smoothness term with $b(\cdot) = \frac{1}{1 + |\nabla I(x)|^2}$ in Eq. (\ref{eq:smooth}). The left ventricle data has no obvious edges, so $b(\cdot) = 1$ is used.

All experiments are performed on 2D slices. The shape prior is randomly taken from the ground truth segmentations. % not sure if we need to mention a future plan for 3D here but maybe better not to
To show the superiority of the proposed integrated FCN-leveset model and its potential for semi-supervised learning we compared five models:
%Following the segmentation procedures were tested on the datasets and their results were compared.
% [dana] you can add some letters or names for those and use them in the table/figures 
\begin{enumerate}[label=(\roman*)]
    \item {\em Pre-trained FCN:} trained with labeled images from the training set (30\% of the liver data; 50\% of the left ventricle data);
    \item {\em Post-processing Level set:} Model (i) with post-processing level set;
    \item {\em Jointly-trained FCN}: the FCN module of the integrated model jointly trained with the level set;
    \item {\em FCN-levelset}: the proposed integrated model as described in Section \ref{sec:fcn-levelset}.
    %Model (iii) with post-processing level set as described in \ref{sec:fcn-levelset}. 
    \item {\em Baseline FCN}: FCN trained on all images in the training set.
\end{enumerate}
The performance of the above methods is evaluated using Dice score and Intersection over Union (IoU) score. Both scores measure the amount of overlap between the  predicted region and the ground truth region. 
%
% Figure 2
\begin{figure}[t]
\centering
\includegraphics[width=.67\textwidth]{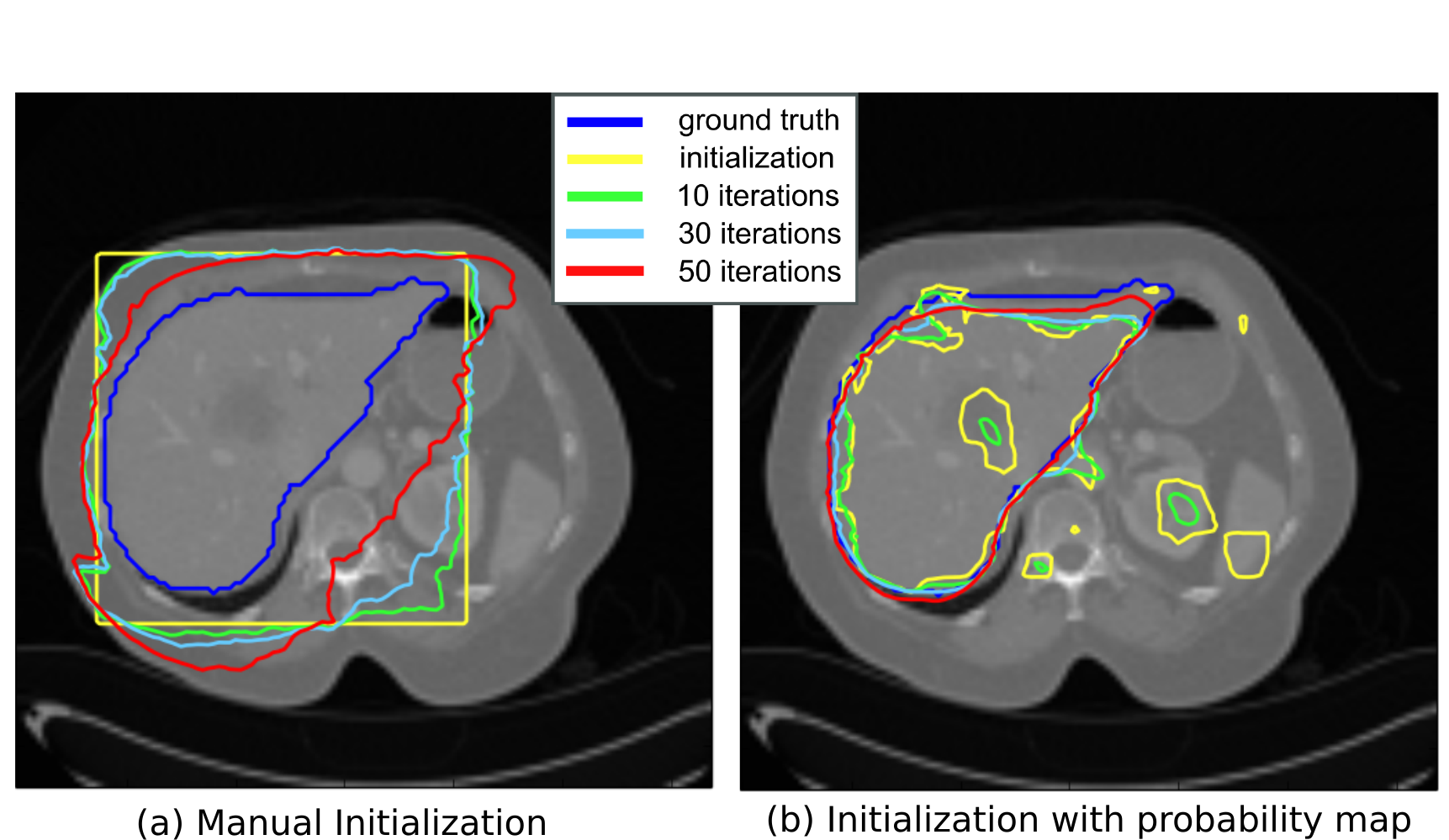}
\caption{Impact of initialization to level set. Both (a) manual initialization and (b) initialized with probability map evolve the contour for 50 iterations. The blue contour is the ground truth; the yellow contour is the initialization and the red one is the segmentation results after 50 iterations.}
%[caption]{\centering Impact of initializations to levelset \hspace{\textwidth}  \small Evolution of the contour in the first 50 iterations. The blue contour is the ground truth; the yellow contour is the initialization from the probability map of the FCN; the red one is the segmentation results after 50 iterations. \normalsize}
%\caption[caption]{\centering Effect of initialization to levelset \hspace{\textwidth}Top: different initialization; Bottom: levelset segmentation after 50 iterations of the levset.}
\label{fig:init}
\end{figure}

%% Figure 3. 
\begin{figure}[b!]
\centering
\includegraphics[width=\textwidth]{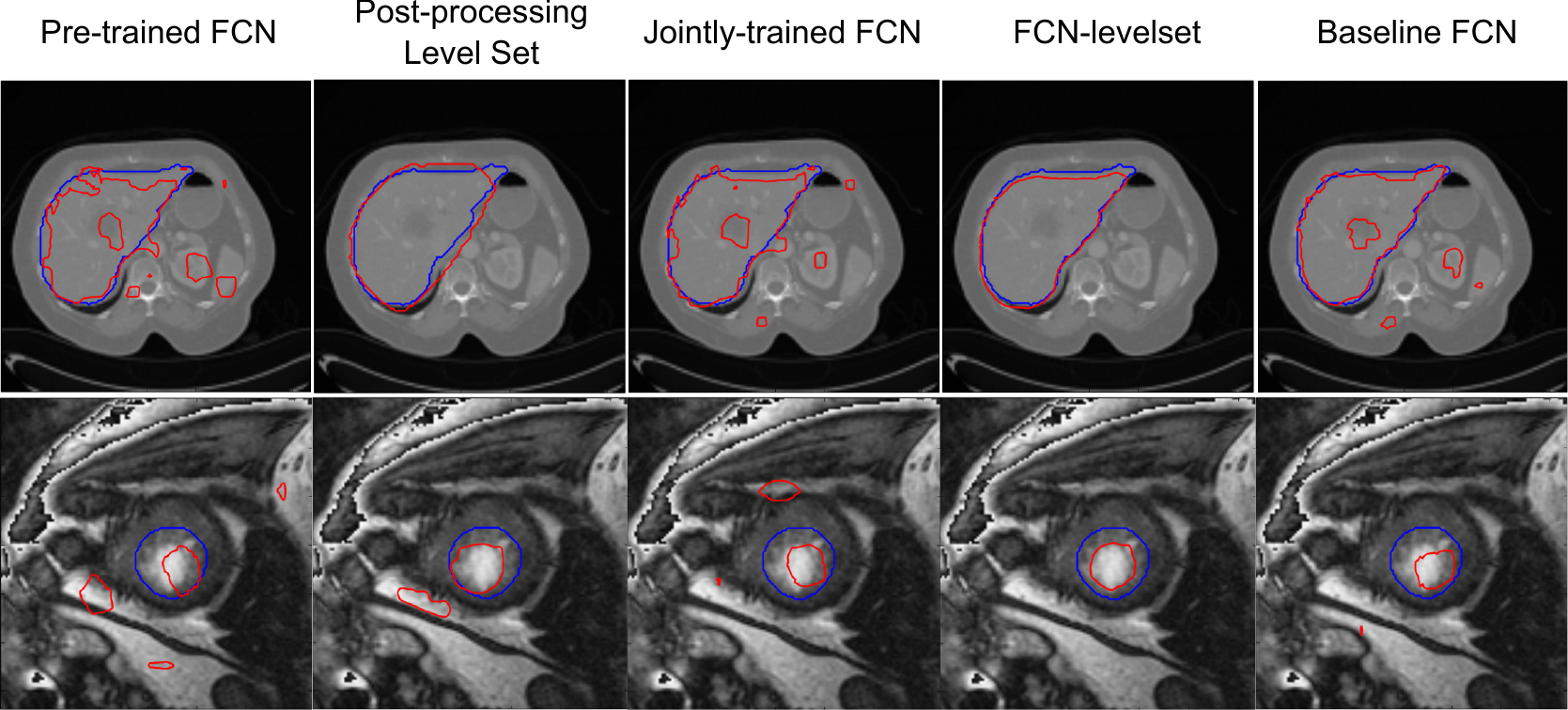}
\caption[caption]{Example segmentation of the five models.
Output contour is in red while the ground truth is in blue. The top and bottom rows show the results on liver dataset and left ventricle dataset, respectively.  The five columns correspond to the five models shown in Table \ref{table:result}.}
\label{fig:comparison}
\end{figure}

\subsection{Results}
To illustrate the sensitivity of the level set method to initializations, we compared the output contours under different initializations, as shown in Fig. \ref{fig:init}. 
The level set with manual initialization Fig. \ref{fig:init}(a) converges to a proper shape (red) but with low precision.
Initialization with the FCN predicted probability map Fig. \ref{fig:init}(b) leads to a faster and more accurate convergence after a few iterations. Notably, when initialized with the probability map, the level set managed to eliminate the wrongly segmented parts (yellow) in the FCN output.

% consider clarifying the way results are presented
 The performance of the five models described in Section \ref{sec:exp} on the two datasets is summarized in Table \ref{table:result} and illustrated in Fig. \ref{fig:comparison}. 
 Both datasets showed that the joint semi-supervised training improves the performance of the deep network.
 %FCN was pre-trained by labeled data which is randomly chosen.
%We randomly chose $30\%$ of the liver data as labeled to train the baseline FCN while $50\%$ was chosen from heart data.
For liver data, the level set initialized with the probability map outperformed the pre-trained FCN by $4\%$. 
%In that case, the prediction from FCN can be refined. 
Through the joint training, FCN was fine-tuned and got an improvement of $3\%$. The performance of FCN-levelset was increased by $8\%$ for Dice Score and $13\%$ for IoU score compared to the pre-trained FCN. Notably, despite using only a few manually labeled images, the integrated model performed even better than the baseline FCN, where the improvement was about $3\%$ in Dice Score and $4\%$ in IoU. 
The integrated FCN-levelset model has clear advantage in datasets where the segmented object presents a prominent shape. 
% ???Min: this belongs to discussion 
\newcommand*{\TitleParbox}[1]{\parbox[c]{2.5cm}{\centering #1}}
\begin{table}[h!]
\centering
\caption{Performance comparison on liver and left ventricle datasets of the five models detailed in Section \ref{sec:exp}
}
% Min: (i) trained with 30\% training data???
\begin{tabular}{llcc}
\hline
Dataset & Model                                     & \TitleParbox{Dice Score}  & IoU \\ \hline
Liver   & Pre-trained FCN         & 0.843       & 0.729 \\
        & Post-processing level set & 0.885       & 0.794 \\
        & Jointly-trained FCN                  & 0.879      & 0.784 \\
        & FCN-levelset                    & \textbf{0.923}      & \textbf{0.857} \\ \cline{2-4}
        & Baseline FCN   & 0.896       & 0.811 \\ \hline \hline
Left Ventricle   & Pre-trained FCN  & 0.754      &  0.605   \\
        & Post-processing level set & 0.678       & 0.620 \\
        & Jointly-trained FCN  & 0.772            & 0.623    \\
        & FCN-levelset & \textbf{0.788}            & \textbf{0.635}    \\ \cline{2-4}
        & Baseline FCN & 0.804      & 0.672 \\ \hline
\end{tabular}
\label{table:result}
\end{table}

Left ventricle data is more challenging for the level set model to distinguish the object from the background. Mainly due to the lack of a clear regional property. However, the joint model still improved the performance of the FCN by $2\%$ during training. The FCN-levelset (trained with only 50\% data) was able to improve the segmentation result by $3\%$ compared to the pre-trained FCN and achieved a comparable performance to the baseline FCN (trained with all data). 

%the performance of FCN was increased by $2\%$ during the joint training. Compared with the Baseline FCN (trained with all data), the integrated model (trained with only 50\% data) still showed a comparable performance.
%the advantage of the integrated model (trained with only 50\% data) was not clear. The main reason is that the left ventricle image doesn't show a clear regional property.  However, the FCN-levelset was still able to improve the segmentation result by $3\%$ compared to the pre-trained FCN.

% Note the improvement in the {\em Fine-tuned FCN} compared to the {\em Baseline FCN} which is due to joint training with the levelset utilizing the unlabeled data. The performance of this {\em Fine-tuned FCN} is comparable with the baseline FCN that is trained on all the data, denoted {\em FCN 100\%}. 

% more comments here 

%% file: files/discussion.tex
%extensions : 3D, make shape pose part of the network, fully integarted system through back-propagation 
In this paper a novel technique for integrating a level set with  a fully convolutional network is presented. This technique combines the generality of convolutional networks with the precision of the level set method. Two advantages of this integration are shown in the paper. First, using the level set initialized with the FCN output can achieve better performance than the FCN alone. Second, as a training technique, by utilizing unlabeled data and jointly training the FCN with the level set, improves the FCN performance. 
%And the performance of the integrated model FCN-levelset achieves even better results. These findings are shown for liver and left ventricle segmentation tasks.
%In the experiments, it has been observed that the amount of improvement that the integrated system can achieve is directly correlated to how well the levelset module performs. In the dataset where there is a strong shape prior
%and clear regional property, levelset shows its advantage and is able to improve the performance of the integrated system. 
%On the other hand, for the data that does not present a clear regional property, level set would not help much.
While the proposed model is only 2D binary segmentation and has a simple shape prior, its extension to 3D and more complex probabilistic shape models is straightforward and we are currently working on it.